\def\year{2020}\relax
\documentclass[letterpaper]{article} % DO NOT CHANGE THIS
\usepackage{aaai20}  % DO NOT CHANGE THIS
\usepackage{times}  % DO NOT CHANGE THIS
\usepackage{helvet} % DO NOT CHANGE THIS
\usepackage{courier}  % DO NOT CHANGE THIS
\usepackage[hyphens]{url}  % DO NOT CHANGE THIS
\usepackage{graphicx} % DO NOT CHANGE THIS
\usepackage{graphicx}  % DO NOT CHANGE THIS
\usepackage{latexsym}
\usepackage{booktabs}
\usepackage{makecell,comment}
\usepackage{tabularx}
\usepackage{verbatimbox}
\usepackage{array}
\usepackage{linguex}
\usepackage{cleveref}
\usepackage{adjustbox}

\usepackage{url}

\newcommand{\citealp}[1]{\citeauthor{#1} (\citeyear{#1})}

\title{What Do You Mean `Why?': Resolving Sluices in Conversations}

\author{Victor Petr\'en Bach Hansen,\textsuperscript{\rm 1 \rm 2} Anders S\o gaard\textsuperscript{\rm 1 \rm 3}\\ 
\textsuperscript{\rm 1}Department of Computer Science, University of Copenhagen, Denmark\\ 
\textsuperscript{\rm 2}Topdanmark A/S, Denmark\\ 
\textsuperscript{\rm 3}Google Research, Berlin\\ 
victor.petren@di.ku.dk, soegaard@di.ku.dk
}

\begin{document}
 
\maketitle
\begin{abstract}
In conversation, we often ask one-word questions such as `Why?' or `Who?'. Such questions are typically easy for humans to answer, but can be hard for computers, because their resolution requires retrieving both the right semantic frames and the right arguments from context. This paper introduces the novel ellipsis resolution task of resolving such one-word questions, referred to as {\it sluices} in linguistics. 
We present a crowd-sourced dataset containing annotations of sluices from over 4,000 dialogues collected from conversational QA datasets, as well as a series of strong baseline architectures.
\end{abstract}

\section{Introduction}
\label{sec:intro}
%Asking short and concise questions, such as `Where?' and `Why?', is a natural way for  humans to seek additional information about a previous passage, resolving the historical context implicitly in an instant, without even having to think about it. Consider the example in Figure \ref{tab:ex1}, when the question {\it When?} is asked, we are effortlessly able to understand that it refers to the time of the bombing that was asked about just previously.
Stand-alone {\em wh}-word questions, such as {\it When?} in Figure \ref{tab:ex1}, are easy for us to understand, but in order to interpret them we need to retrieve implicit information from context. Learning to do so is an instance of {\em sluicing}, an ellipsis phenomenon, defined by Ross (\citeyear{Ross:69}) as `the effect of deleting everything but the preposed constituent of an embedded question, under the condition that the remainder of the question is identical to some other part of the sentence, or a preceding sentence.' In the context of conversations, one-word {\em wh}-word questions are particularly frequent \cite{anand2016,ronning-etal-2018-sluice}, and because they are often hard to resolve, they seem to be a frequent source of error in conversational question answering \cite{quac,coqa} and dialogue understanding \cite{Vlachos:Clark:14}. 
We refer to this type of sluicing as {\em conversational sluicing}. 

\begin{figure}[h]
    \centering
    %\begin{tabularx}{9cm}{m{0.8cm} l}
    \begin{tabularx}{\linewidth}{ll}
        $Q_1$:          & {Where was the bombing?} \\
        \adjustbox{valign=b}{$A_1$:} & {\makecell[l]{San Diego's Edward J. Schwartz Federal\\Courthouse.}} \\
        $Q_2$: & {When?} \\
        \cmidrule(lr{1em}){2-2}
        $R_1$: &  When [{\it was the bombing}?] \\
        \adjustbox{valign=t}{$R_2$:} &  \makecell[l]{When [{\it was the bombing of San Diego's}\\ {\it Edward J. Schwartz Federal Courthouse}]?} 
    \end{tabularx}
    \caption{Example of conversational sluicing. $Q_1$ and $A_1$ provides a context for the second question $Q_2$ which has multiple correct resolutions, denoted in brackets, such as $R_1$ and $R_2$.}
    \label{tab:ex1}
\end{figure}

%Answering questions with omitted, or {\em elided}, information is a challenging task for machines to resolve. This is one of the reasons that conversational Question Answering (QA) datasets, such as Question Answering in Context (QuAC) \cite{quac} and Conversational Question Answering (CoQA) \cite{coqa} are are harder to perform well on, compared to their non-conversational counterparts, e.g. the Stanford Question Answering Dataset (SQuAD) \cite{Rajpurkar-squad, Rajpurkar-squad2}. \newcite{anand2016} argue that, especially question-answering systems have to learn to effectively deal with ellipsis, due to the fact that both elliptical questions and answers are very common in discourse, in order to produce good answers. \newcite{Nielsen03acorpus-based} identifies three subtasks that are essential for successfully handling ellipsis: detecting ellipsis, identifying antecedents and ellipsis resolution. In this work we tackle the last task, namely resolving instances of conversational sluices. % correctly interpreting the sluice.% \ldots

%The goal of the CoQA challenge is to measure the ability of machines to understand a text passage and answer a series of interconnected questions that appear in a conversation. 

Unlike previous work where sluice resolution is treated as predicting the span of the antecedent \cite{anand2016,ronning-etal-2018-sluice}, we frame conversational sluice resolution as a Natural Language Generation (NLG) task, in which we seek to automatically generate the full question, given a question-answer context and a one-word question. To this end, we provide a novel corpus of conversational sluice annotations and explore a series of strong baselines and their performance on this dataset.

\paragraph{Contributions} In this paper we introduce the task of resolving conversational sluicing, a pervasive and challenging ellipsis phenomenon. %, i.e. learning how to extract elided information from one-word questions when given a previous context, using NLG. 
We crowd-source a new dataset containing over 4000 annotated sluices, gathered from existing conversational QA datasets. We conduct a series of baseline experiments on this task, using both encoder-decoder frameworks, as well as language modelling objectives, and show through human evaluation of the predicted resolutions that these baselines are quite strong and at times even rival the quality of human annotators.

\section{Background}
\paragraph{Sluicing}
\label{sec:background}
Ellipsis is the linguistic phenomenon that describes the omission of one or more words from a phrase that can be retrieved from a previous context. Sluicing is a case of ellipsis where content is elided from a question, leaving behind only the {\em wh}-remnant. 
\citealp{anand2016} and \citealp{ronning-etal-2018-sluice} consider two types of sluices, namely {\em embedded} sluices and {\em root} sluices, also sometimes referred to as {\em bare} sluices. 
%R{\o}nning et al.~ (\citeyear{ronning-etal-2018-sluice})

\ex. \label{ex:1} My neighbor said he would stop by, but I don't know when [{\it he would stop by}].

\ex.  \label{ex:2} \a. My neighbor is stopping by.
     \b. When [{\it is the stopping by}]?

\begin{comment}
\addvbuffer[8pt 8pt]{\begin{tabular}{rp{0.8\linewidth}}
    (1)&  My neighbor said he would stop by, but I don't know when [{\it he would stop by}].\label{ex:1}\\
    (2)&  \begin{tabular}{ll}\label{ex:2}
        A: & My Neighbor is stopping by.  \\
        B: & When [{\it is the stopping by}]? 
    \end{tabular}
\end{tabular}}
\end{comment}

In Example (1), we see an instance of embedded sluicing where the question is a part of a larger structure, and \ref{ex:2} is an example of a root sluice where the {\it wh}-fronted ellipsis is an utterance in itself, i.e. in a root environment. \citealp{anand2016} note that sluicing in dialogue often differs from sluicing in single-authored text, with root sluices being more prevalent in dialogue. In dialogue, using sluices -- and ellipsis in general -- requires a level of mutual understanding.  \citealp{colman-etal-2008-quantifying} therefore use ellipsis in dialogue as a means of quantifying mutual understanding in conversations. 

\citealp{fernandez-etal-2007-classifying} focus on the task of classifying occurrences of single-word sluices in conversations and call these {\em bare} sluices. They categorize such sluices into distinct categories; (i) {\it direct}, which is the case where the sluice queries for additional information that was quantified, either explicitly or implicitly, in the previous utterance; (ii) {\it reprise}, where the speaker is unable to understand an aspect of the previous utterance, which the initial speaker assumed as presupposed; (iii) {\em clarification}, where the speaker uses the sluice to ask for clarification of the entire preceding utterance; (iv) {\em Wh-anaphor}, where the antecedent is a wh-phrase; and (v) {\em unclear}, the case where it is difficult understand what the sluice conveys, usually because of a lack of proper context. Note that the direct, reprise and clarification sluices are relatively easier to resolve, since their answer can always be retrieved from the previous sentence. Our corpus therefore ignores the first three types of conversational sluices and focuses on (bare or stand-alone) {\em wh}-anaphors; in our annotation experiments below, we also allow annotators to skip unclear instances. Similarly, \citealp{baird-etal-2018-classifying} presented classification experiments learning to distinguish between different types of sluices in dialogue. 

Conversational sluices usually depend on their  question-answer context, and can span both the previous utterances, i.e. the answer, as well as the previous question, whereas direct/reprise/clarification sluices only requires retrieval of context from the previous utterance.
Consider the multi-turn example: 

\begin{center}
\begin{tabular}{rp{0.5\linewidth}}
    A:&  Did Ned have family?\label{ex:1}\\
    B:&  Yes.\\
    A:&Who [{\it was Ned's family}]?
    %\begin{tabular}{ll}\label{ex:2}
    %    A: & My Neighbor is stopping by.  \\
    %    B: & When [{\it is the stopping by}]? 
    %\end{tabular}
\end{tabular}
\end{center}

Resolving this sluice, depends on both the question initially asked by speaker A in addition to the outcome of the answer from speaker B. Looking only at the previous utterance, in this case, would not provide sufficient context, as the {\it Yes/No} utterance of speaker B determines what information from speaker A is relevant for the resolution.

The first efforts to resolve (non-conversational, standard) sluices, by identifying the antecedent of the {\em wh}-remnant, is due to \citealp{Anand2015}, who describe a linguistically-informed annotation scheme for resolving sluices. They present a dataset of ~3.100 annotated examples of sluices extracted from the New York Times section of the English Gigaword corpus.
\citealp{anand2016} presented the first sluice resolution system, achieving decent performance, but \citealp{ronning-etal-2018-sluice} subsequently presented a neural multi-task architecture outperforming their original model by some margin.

A few researchers have explored ellipsis resolution in dialogue: \citealp{Kazuhide:1998} discussed the importance of being able to resolve sluices to understand dialogue. They showed that for certain types of conversational ellipsis, it is possible to achieve good results with simple classification algorithms. Their results are not comparable to other results in the literature, because they focus on a small subset of phenomena, rely on linguistic preprocessing, and consider ellipsis phenomena in Japanese. \citealp{ronning-etal-2018-sluice} also evaluates on conversational data from English Open Subtitles. Their results suggest that resolving sluices in dialogue is harder than domains such as newswire, with $F_1$ resolution scores dropping from $>0.7$ in newswire to around $0.5$ for conversations. As stated, these previous approaches to sluice resolution differs from ours, as we seek to generate a reconstruction of the sluice, not predict the span of the antecedent. Due to the fact that in a conversational context, the antecedent is conditioned on the response to the initial question in our question-answer context, it often results in disjoint antecedent spans, which cannot be represented in the architecture proposed by \citealp{ronning-etal-2018-sluice}. The advantage of resolving the sluice using NLG approaches is that for most downstream purposes, a fluent paraphrase of the {\it wh}-word and the antecedents is preferred and not only an antecedent span, that as stated above, can be non-coherent.

\paragraph{Question Generation}
Researchers have worked on question generation from text paragraphs \cite{Zhao:ea:18}, relative clauses \cite{Khullar:ea:18}, 
SQL queries \cite{Guo:ea:18}, 
knowledge bases \cite{Serban:ea:16}, etc. \citealp{Khullar:ea:18}, which is probably the problem set-up most similar to ours, albeit much simpler, consider relative clauses such as in {\it I am giving fur balls to John who likes cats}. Their simple observation is that relative clauses translate almost straight-forwardly into questions, e.g., {\it Who likes cats?}. Using a small set of heuristic rules, they extract relative clauses and use them to generate training data for machine comprehension. Our task is considerably harder, since we deal with an ellipsis phenomenon that requires us to find antecedents in the previous dialogue turns. Our approach is also very different. While \citealp{Khullar:ea:18} can solve their problem with simple rules, we cannot, and we therefore present neural baseline architectures originally developed for language modeling and transduction tasks.

%https://aclweb.org/anthology/D17-1090
%https://www.aclweb.org/anthology/D18-1424
%https://aclweb.org/anthology/P18-3022
%https://www.aclweb.org/anthology/N18-2090

\section{A Conversational Sluicing Dataset}
In this work, we present a crowd-sourced annotated sluicing dataset. The dataset consists of sluice occurrences 
in conversational question answering contexts. The conversations are teacher-student dialogues, where the teacher asks questions about a background text passage, and the student has to answer the teacher's questions. 
Sluices, and ellipsis in general, are frequent in the data. Each datapoint consists of (i) an initial question, $Q_1$, (ii) an answer to $Q_1$, $A_1$, together forming the QA context $(Q_1,A_1)$, (iii) a one word follow-up {\it wh}-question, $Q_2$, (iv) a gold annotated resolution, $R$, to the sluice in (iii), written in free-text. The resolutions are what we crowd-source to construct the new conversational sluicing dataset. Given question-answer context pairs $(Q_1, A_1)$ and one-word follow-up questions $Q_2$, we seek to resolve conversational sluices by generating the full questions $R$ by explicitly generating the elided context, therefore framing it like a NLG task, rather than an antecedent selection task as done by \citealp{ronning-etal-2018-sluice} and \citealp{anand2016}. This also dramatically simplifies the annotation process as we only seek a resolved sluice in the form of $R$ instead of the annotation scheme used by \citealp{Anand2015} and \citealp{ronning-etal-2018-sluice}, i.e. explicitly annotating the antecedent, sluiced expression, main predicate of the antecedent clause as well as potential correlates in addition to annotations for the auxiliary tasks.

This section describes the process of collecting and cleaning the annotations, and presents a quantitative and qualitative analysis of the dataset.

\paragraph{Data Collection Methodology} 
In order to obtain our conversational sluicing dataset, we crawl existing conversational QA datasets, namely QuAC\footnote{\url{https://quac.ai/}} and CoQA,\footnote{\url{https://stanfordnlp.github.io/coqa/}} for question-answer contexts with one-word follow-up questions. Specifically, we identify all occurrences of five one-word questions: \textit{Why?}, \textit{What?}, \textit{Where?}, \textit{Who?} and \textit{When?}. For each such question, we construct a tuple of the previous QA context and the follow-up question. This process results in roughly ~4200 examples of conversational sluices.

We then proceeded to ask Amazon Mechanical Turkers (AMT) to fill out the remainder of the question as asked by the interrogator based on the the question-answer context pair. 
In order to not impose too many restrictions on the annotators, we left it up to the AMT workers to decide how much of the elided information they wanted to include in their answer, as a conversational sluice can often be solved in multiple ways. For example, in Figure \ref{tab:ex1} we consider both $R_1$ and $R_2$ as correct resolutions to the conversational sluice, even if $R_1$ did not specify the PPN \textit{San Diego’s Edward J. Schwartz Federal Courthouse} as the location of the bombing. In general, annotations often differed in whether modifiers and relative clauses were included, whether or not previous anaphora was resolved, etc. If the previous question and answer did not provide enough context to fill out the elided information, the workers were informed to simply skip it and move on to the next example. We collected a single annotation for each sluice in the training and test splits, and three annotations for each sluice in the test set. For the test set, we use each unique annotation as a separate datapoint.
We allocated 1 minute per annotation and paid the workers $\$ 0.13$ for each accepted annotation. The average time spent per assignment was around $20$ seconds, which results in an hourly rate of $\$23.4$. The total cost of the crows-sourcing process was $\$ 797$.

For our final corpus, we filter out the examples skipped by the annotators, in addition to the conversational sluices whose $Q_1$ context is less than 3 words, as these showed empirically to not contain enough information, usually due to $Q_1$ being a sluice itself. Consider, for example: 

\begin{center}
\begin{tabular}{rl}
        $Q_1$:          & {By who?} \\
        $A_1$:          & {Unknown assailants.} \\
        $Q_2$:          & {Where?}
    \end{tabular}
\end{center}

\noindent Without first resolving the sluice \textit{By who?}, we are unable to properly identify the antecedent, as it is unclear whether or not $Q_2$ refers to the current location of the assailant or the location of the actual assault. These are also the sluices categorized as {\it Unclear} by \citealp{fernandez-etal-2007-classifying}. After cleaning, we reduced the initial size from 4980 to 4175 datapoints.

\paragraph{Corpus Statistics} 
In Table \ref{tab:stats}, we show the distribution of the different {\it wh}-questions across the various splits in our corpus.
The dataset contains both instances of conversational sluices as well as reprise/direct/clarification sluices. We release the raw annotated version of the conversational sluicing corpus, as well as our cleaned version which we report our results on, including the splits used.\footnote{ \url{https://github.com/vpetren/conv_sluice_resolution}}
\begin{table}[h]
    \centering
    %\small
    \resizebox{.95\columnwidth}{!}{
    \begin{tabular}{c|ccccc|c}
        \toprule
        Split & {\it Why} & {\it Where} & {\it Who} & {\it What} & {\it When} & Total  \\
        \midrule
        train   & 851 & 714 & 513 & 302 & 702 & 3082 \\ 
        val     &  84 & 71  & 54  & 39  & 52  & 300\\ 
        test    & 229 & 183 & 97  & 83  & 201 & 793\\
        \midrule
        Total   & 1164& 968 & 664 & 424 & 955 & 4175 \\
        \bottomrule
    \end{tabular}
    }
    \caption{Statistics of the {\em wh}-word distribution across the different splits for our conversational sluicing dataset.}
    \label{tab:stats}
\end{table}

Empirically, we did not observe many long distance dependencies between the sluice and corresponding antecedent, as it was found within a three-turn window a majority of the time (around 95$\%$). \citealp{ronning-etal-2018-sluice} similarly reports that long term dependencies (3 or more sentences between sluice and antecedent) are very rare (around 1$\%$). Solving these rare dependencies would also be an interesting task, but is however outside the scope of this work. This dataset provides a reasonable limitation for a stab at an already challenging phenomenon. 

\paragraph{Performance Metrics} Natural language generation systems are often evaluated in terms of BLEU scores \cite{Papineni:ea:02} and on subsamples of standard corpora. Neither are likely to be optimal. Finding an appropriate performance metric that correlates with human judgments of resolution quality, is crucial to ensure progress on conversational sluicing resolution; and evaluating across different samples is equally important to avoid community-wide over-fitting to one particular sample. We hope to be able to contribute to improving both performance metrics and the data situation, but for now we also report the performance of our baseline systems in terms of BLEU scores on a random subsample. In order to combat the bias introduced by BLEU, we supplement the scores with alternative performance metrics, as well as with human judgments from professional annotators.
BLEU originally was intended for corpus-level evaluation and has several limitations when applied at the sentence-level \cite{Rapp:09}. We therefore also include the GLEU metric, as proposed by \cite{Wu16GNMT}, which according to their experiments, is better suited for sentence-level evaluation, while still correlating well with BLEU on the corpus-level.\footnote{We use the sentence-level GLEU and BLEU implementations provided by NLTK with the smoothing function introduced by \citealp{Lin:2004}}
In addition to BLEU and GLEU we also measure the the character $n$-gram F-score ({\sc chrF}) \cite{popovic-2015-chrf}, as well as the precision ({\sc chrP}) and recall ({\sc chrR}). We use $\beta=3$, i.e. assigning a higher weight to recall, as it has been shown to correlate better with human judgements than other popular automatic machine translation metrics, such as BLEU and ROGUE-L. For $n$ we use $4$-grams.

Given the shortcomings of automatic evaluation metrics, we also include a human evaluation study. We sample $n$ contexts along with the gold sluice resolution and the resolutions generated by our baseline models from the test set and ask human evaluators to rank them according to relative quality. We obtained judgments of 100 document instances and report on these experiments in \S \ref{sec:analysis}. 

\begin{table}
  \centering
  %\small
  \resizebox{.95\columnwidth}{!}{
    \begin{tabular}{l||r|r|r|r|r}
    %\toprule
    %\multicolumn{1}{c}{}  & \multicolumn{5}{c}{\sc Automatic} & \multicolumn{2}{c}{\sc Human}\\
    %\midrule
    %\cmidrule{2-5}
    \toprule
    Model               & {\sc GLEU}        & {\sc BLEU}        & {\sc chrF}        & {\sc chrP}        & {\sc chrR}    \\
    \hline
    {\sc C\&E Q1}       &  0.035            & 0.043             & 0.114             & 0.034             & 0.166         \\
    {\sc C\&E A}        &  0.010            & 0.016             & 0.034             & 0.011             & 0.048         \\
    {\sc LSTM-seq2seq}  &  0.232            & 0.304             & 0.276             & 0.311             & 0.274         \\
    {\sc Transformer}   & 0.337             & \textbf{0.391}    & 0.443             & 0.461             & 0.442         \\
    {\sc GPT-2}         & 0.067             & 0.117             & 0.138             & 0.109             & 0.167         \\
    {\sc GPT-2 (FT)}    & \textbf{0.348}    & \textbf{0.391}    & \textbf{0.467}    & \textbf{0.499}    & \textbf{0.470}\\
    \midrule
    {\sc Ann Agree}      & 0.570             & 0.589             & 0.712             & 0.704             & 0.720         \\
    \bottomrule
    \end{tabular}%
  }
  \caption{Results on our conversational sluicing dataset for a series of baseline architectures. We measure the performance using BLEU, GLEU and character $n$-gram F-score, precision and recall on the test split. In the last row, {\sc Ann Agree} denotes the inter-annotator agreement as the average between two randomly sampled gold annotations from each data point of the test set.
  }
  \label{tab:results}%
\end{table}%

\paragraph{Annotation Quality} In the last row of Table \ref{tab:results}, {\sc Ann Agree}, we report the inter-annotator agreement scores of the test set. For each of the 3 collected annotation per conversational sluice instance, we sample 2 of them and calculate BLEU, GLEU, chrF, chrP and chrR scores between them as a measurement of annotator agreement. As different annotations can be considered correct sluice resolutions, we use this measurement as a means to set an expectation for the performance ceiling of our models. In general we observe that there seems to be reasonably high annotator agreement scores compared to the best performing models, but still indicates that the sluices can be solved in multiple correct ways.

\begin{table}
    \centering
    \begin{tabular}{l||r|r}
       % \multicolumn{1}{c}{}  & \multicolumn{5}{c}{\sc Automatic} & \multicolumn{2}{c}{\sc Human}\\
    %\midrule
    %\cmidrule{2-5}
    \toprule
    Model               & {\sc MRR }    & {\sc $r_1$} \\
    \hline
    {\sc LSTM-seq2seq}  & 0.295         &   0.005\\
    {\sc Transformer}   & 0.381         &   0.030\\
    {\sc GPT-2 (FT)}    & {\bf 0.529}   &   {\bf 0.190}\\
    \midrule
    {\sc Gold}      & { 0.879} &   { 0.775}\\
    \bottomrule
    \end{tabular}
    \caption{The results of the human judgement experiment. To obtain human judgments, we asked three annotators to rank the output of three systems and the crowd-sourced gold annotations. MRR is the mean reciprocal ranking, and $r_1$ refers to the fraction of presented examples where the model was ranked as number 1. Our results show that the fine-tuned GPT-2 model produces favorable resolutions, both in terms of automatic as well as human evaluation and 1/5 instances {\em better} than gold annotations.}
    \label{tab:human}
\end{table}

\section{Experiments}
%This section presents the experimental setup, our baseline architectures, and the results obtained in our experiments. 
In our experiments, we use the splits outlined in Table \ref{tab:stats} (also made publicly available). We preprocess our data by appending the QA context and one-word question together, converting the input sequence into the format \texttt{ <s> Q1 <del> A1 <del> Q2 </s>} and the target sequence we seek to generate as \texttt{<s> R </s>}. Here \texttt{<del>} is a special delimiter token, and \texttt{<s>} and \texttt{</s>}, denote the beginning and end of the sequence. In addition to this, we only preprocess the data by performing lower-casing and tokenization.
%We train all our models using cross-entropy until loss convergence.
\subsection{Baseline models}
In this section, we present a number of different baseline architectures and heuristics for the task of conversational sluice resolution.

\paragraph{Copy \& Edit Heuristics} Seeing as the structure of the resolved sluice in some cases takes on the form of either $Q_1$, especially in the cases where a yes/no answer precedes it, or $A$, as seen in Figure \ref{tab:ex1}, we propose two simple copy and edit heuristics. (i) Given the QA-context and our conversational sluice $Q_2$, we simply replace the wh-question word in $Q_1$ with $Q_2$ and use this augmented question as the resolution to our sluice. We refer to this as {\sc C\&E Q1}. (ii) Similarly, we can copy the answer from $A$ and prepend the $Q_2$ sluice to it. We refer to this as {\sc C\&E A}.

\paragraph{LSTM-seq2seq} Sequence-to-sequence models \cite{sutskever14}, or seq2seq, have previously been successfully applied to conversational modelling tasks \cite{VinyalsL15}. They use the encoder-decoder framework, where an input context is encoded by an encoder-module, usually a variant of Recurrent Neural Networks (RNNs), and decoded by a decoder-module, into the target sequence. For both the encoder and decoder, we use a standard two-layer LSTM \cite{Hochreiter:1997}, with a hidden state size of 512, and regularized using a dropout rate of $0.5$. 
We initialize the embedding matrix with 300 dimensional GloVE \cite{pennington2014glove}, which remains fixed during training. We optimize the end-to-end network using Adam \cite{KingmaB14}, with the default learning rate of $0.001$.\footnote{Implementation is based on \url{https://github.com/bentrevett/pytorch-seq2seq}.}

\paragraph{Transformer} The transformer architecture \cite{vaswani17} is now the {\it de facto} standard architecture in machine translation and has paved the way for state-of-the-art pre-trained contextual language encoders such as BERT \cite{bert} and the OpenAI GPT-2 \cite{radford2019language}. While still adopting the encoder-decoder framework, instead of processing the source and target sequences sequentially, it relies on a multi-headed self-attention mechanism, attending over the entire sequence at same time, allowing for greater parallelization and a positional encoding of the sequence, ensures that contextual information is maintained.
As our conversational sluicing resolution corpus is small in comparison to the corpora used in the experiments by \citealp{vaswani17}, we limit ourselves to three encoder/decoder layers to 3 (compared to 6 in their work), after observing improvements on our validation data.\footnote{Implementation is based on \url{https://github.com/jadore801120/attention-is-all-you-need-pytorch/}} As with the LSTM-seq2seq model, we initialize the embedding matrix with 300 dimensional GloVE embeddings, but otherwise we use the defualt hyperparameters.

\paragraph{GPT-2} The Generative Pretrained Transformer-2 (GPT-2) \cite{radford2019language}, trained to simply predict the next word in 40GB of Internet text, has since its introduction been used to generate state-of-the-art performance on multiple language modelling datasets. The GPT-2 architecture, as mentioned above, is based on the transformer architecture. In our experiments, we use the small pretrained model released by OpenAI (117M parameters). We experiment both with the pretrained GPT-2 model as is, as well as with fine-tuning it on our sluicing corpus.
%\footnote{Fine-tuning the GPT-2 model builds on the implementation by \url{https://github.com/nshepperd/gpt-2}} 
When fine-tuning the model, we simply concatenate the input and output sequences together and input them to the language model. Unlike the LSTM-seq2seq and Transformer, we do not fine-tune the GPT-2 model until convergence, but instead we ran it for 18 hours on an Nvidia TitanX GPU. We also report the performance of the GPT-2 model on our task when no fine-tuning has taken place.

\paragraph{Other baselines considered}
Inspired by \citealp{hill16} and \citealp{lample17}, we also experimented with pretraining the {\sc Seq2Seq-LSTM} and {\sc Transformer} architectures with sequential de-noising autoencoder objectives. We collected a dataset consisting of ~350.000 questions from CoQA, QuAC and SQuAD 2.0, making sure not to include cases of sluices, hypothesizing that this would allow the encoder and decoder to learn the internal structure and representation of questions. After pre-training, we fine-tune the architectures on our conversational sluicing data. These experiments did, however, not lead to any improvements in the performance when using automatic metrics. A manual inspection of the generated resolutions did not reveal any noticeable improvements over their non pre-trained counterparts, so we do not report the results below.

Again, we stress that due to the reasons listed above, i.e. incompatible annotation schemes between our work and that of \citealp{ronning-etal-2018-sluice} as well as the lack of flexibility that a span-prediction model provides, we do not use their work as a baseline. We hypothesize that our heuristics, {\sc C\&E Q1} and {\sc C\&E A}, will serve as an indication as to what we can expect from these types of models.

\subsection{Results}
Table \ref{tab:results} summarizes the results from our baseline models on our conversational sluicing corpus, using standard automatic performance metrics. 
The results suggest that the fine-tuned {\sc GPT-2} architecture is superior to all other baselines across the board, achieving scores closest to the inter-annotator ceiling, with the {\sc Transformer} model rivalling it on the BLEU score. 
Although the {\sc C\&E Q1} and {\sc C\&E A} heuristics could seem like strong baselines, as some of the examples in Table \ref{tab:model_pred} and Figure \ref{tab:ex1} might suggest, our results tells a different story. Again, this illustrates the flexibility that is required to resolve these conversational sluices, which a non-disjoint antecedent span fails to capture.
We can observe that without the task-specific fine-tuning, the GPT-2 model falls short, as it ultimately just proceeds to generate what comes after the sluice, not resolving it. However, this extensive pretraining does shine through compared to the {\sc Transformer} model, when fine-tuned on our dataset as we also can see from our human evaluation (illustrated in \ref{tab:human}), which we discuss in the next section.

\begin{figure*}[h]
    \centering
    \includegraphics[width=.95\textwidth]{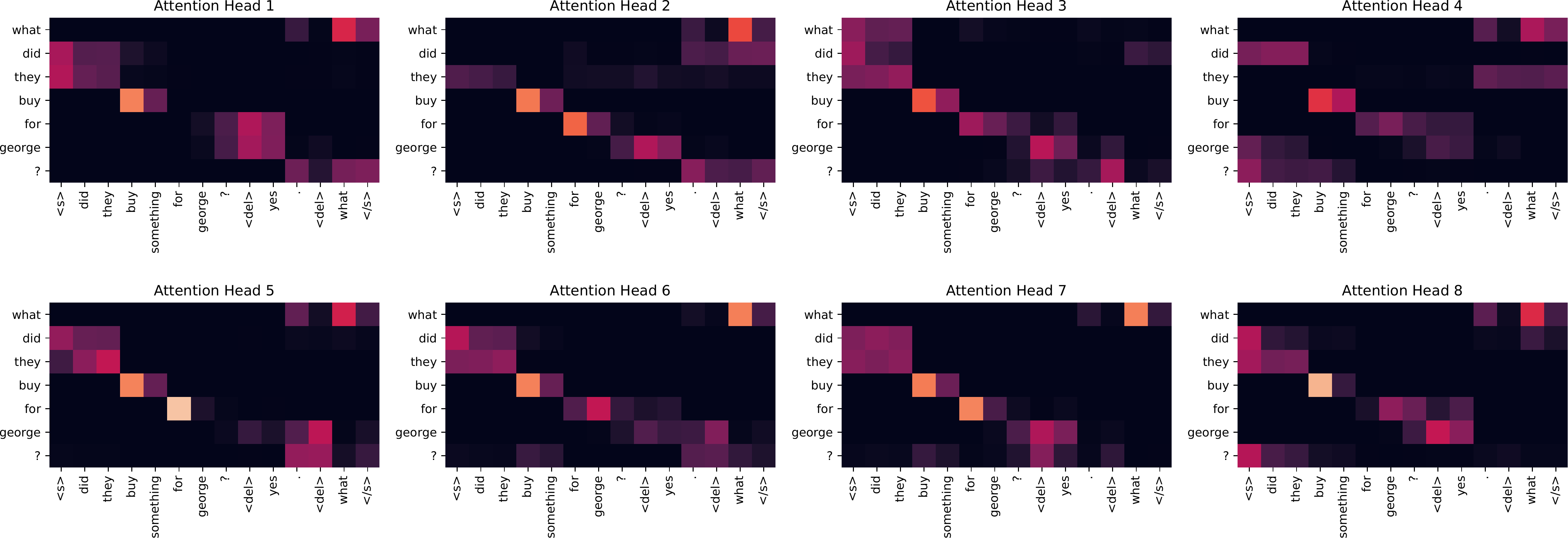}
    \caption{Illustration of the attention weights from all the 8 attention heads in the final decoder layer of the Transformer network. The x-axis corresponds to the position in the input sequence, whereas the y-axis corresponds to the output sequence.}
    \label{fig:attn_head}
\end{figure*}

\begin{figure*}[h]
    \centering
    \includegraphics[width=.95\textwidth]{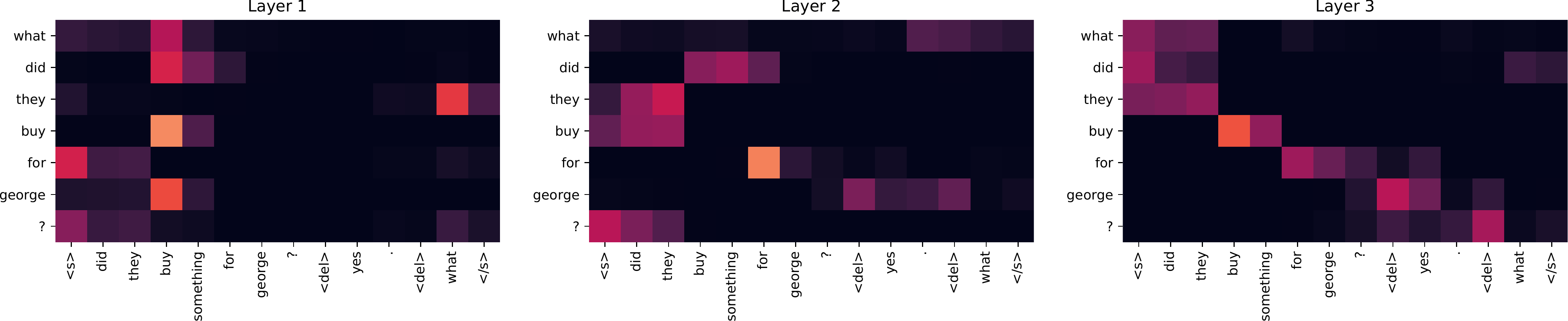}
    \caption{Illustration of the attention weights from a single attention head in the 3-layer Transformer network, during decoding. The x-axis corresponds to the position in the input sequence, whereas the y-axis corresponds to the output sequence.}
    \label{fig:attn_layer}
\end{figure*}

\begin{table*}[h]
    \centering
    \small
    \begin{tabular}{p{0.21\textwidth}|p{0.16\textwidth}|p{0.16\textwidth}|p{0.16\textwidth}|p{0.17\textwidth}}
        \toprule
        {\sc Context} & {\sc LSTM-s2s} & {\sc Transformer} & {\sc GPT-2} & {\sc Gold} \\
        \hline
        \makecell[l]{
            \begin{tabular}{rl}
                $Q_1$: &  \makecell[l]{What did Susie do?} \\
                $A_1$: & Woke up. \\
                $Q_2$: & When?
        \end{tabular}} &  When did they go?? & When did Susie woke up? & When did Susie wake up? & When did Susie wake up?\\
        \hline
        \makecell[l]{
            \begin{tabular}{rl}
                $Q_1$: &  \makecell[l]{Did the island ever \\change its form of\\ government?} \\
                $A_1$: & Yes. \\
                $Q_2$: & When?
        \end{tabular}} &  When did the objective of the?? Scotland? & When did the island change its form of government? & When did the island form? & When did the island change its form of government? \\
        \hline
        \makecell[l]{
            \begin{tabular}{rl}
                $Q_1$: &  \makecell[l]{Is there any \\mysterious \\character?} \\
                $A_1$: &  Yes.\\
                $Q_2$: &  Who?
            \end{tabular}} & Who is the other?? & Who are the character in? & Who was the famous person that was added to the story? & Who is the mysterious character?\\
        \hline
        \makecell[l]{
            \begin{tabular}{rl}
                $Q_1$: &  \makecell[l]{Did he say any-\\thing before \\leaving?} \\
                $A_1$: &  Yes.\\
                $Q_2$: &  What?
            \end{tabular}} & What did he do ? & What did he say? & What did he say before he left? & What did he say?\\
        \bottomrule
    \end{tabular}
    \caption{Generated output from our series of baselines, given a question-answer context, ($Q_1$, $A_1$) and follow-up one-word question. Examples are taken from the test split.}
    \label{tab:model_pred}
\end{table*}

\section{Analysis}
\label{sec:analysis}
\paragraph{Human judgment of generated resolutions}
Knowing that our automatic evaluation metrics can be biased when applied at the sentence-level, we also include a human evaluation study on a random sample of 100 instances of sluices. We asked human evaluators to rank the resolutions generated by our best performing models, i.e. the {\sc LSTM-seq2seq} architecture, the {\sc Transformer} architecture, our fine-tuned GPT-2 model, as well as the human annotators' resolutions, by their quality and relevance in a QA context. We presented the four resolutions in random order and asked subjects to place them, from best to worst. If they deemed two or more candidates to be equally good or bad, we instructed them to simply order these randomly. We report performance using the Mean Reciprocal Rank (MRR), and what we refer to as $r_1$, which denotes the fraction of presented examples where the model was ranked as number 1.
Our evaluation, shown in Table \ref{tab:human}, reveals that the human judges tend to favour the resolutions provided by {\sc GPT-2 (FT)} over the ones produced by the {\sc Transformer}~architecture. In fact, the GPT-2 resolutions are chosen over all other resolutions, including our gold standard, in 1/5 instances. 
Generally we see the same trend in the human evaluation experiment as with the automatic metrics, except that the GPT-2 model now significantly outperforms the other baselines. We believe this can be attributed to the fact that our human judges may be biased toward selecting well-formed resolutions, and the GPT-2 language model may simply be better at generating fluent language.

To illustrate an instance where GPT-2 can generate a more expressive resolution than our gold standard, consider the example in Figure \ref{tab:ex_hum}. Here, the fine-tuned OpenAI GPT-2 model generates a resolution that the judges found to be better than the gold standard, not because the gold-standard was wrong, but because the automatic resolution was more informative, easing interpretation. 
\begin{figure}[h]
    \centering
    \begin{tabularx}{9cm}{rl}
        $Q_1$:          & {\makecell[l]{Is anyone who works with them mentioned?}} \\
        $A_1$:          & {Yes.} \\
        $Q_2$:          & {Who?} \\
        \cmidrule(lr{1em}){2-2}
        $R_{GPT2}$:     & {Who [else is mentioned]?} \\
        $R_{Gold}$:     & {Who [is mentioned]?} \\
    \end{tabularx}
    \caption{Conversational sluice resolution by the fine-tuned GPT-2 model that is judged better than the gold standard by our annotators.}
    %\vspace{-3mm}
    \label{tab:ex_hum}
\end{figure}

\paragraph{Visualization of attention weights}
An advantage of the attention mechanism, is that it allows for high interpretability, when it comes to the showing where in the input sequence the model is attending at a given time-step. To get a better understanding of where the Transformer attends during decoding, we visualize the internal attention mechanisms of the model trained on our conversational sluicing corpus. Figure \ref{fig:attn_layer} shows the attention matrix heatmaps of a single attention-head in each layer and Figure \ref{fig:attn_head} shows the attention matrix heatmaps for each of the 8 attention-heads in the last layer of the Transformer. When looking at Figure \ref{fig:attn_layer}, we see that the various layers encode different levels of information, with the attention-head of the last layer seemingly being the most structured.
From Figure \ref{fig:attn_head}, we can observe that the various attention-heads mostly present the same pattern. When generating the first word of the resolution, the attention is at the end of the input sequence, i.e. on the {\it wh}-fronted ellipsis. Generating the subsequent tokens then shifts the attention back to the beginning of the input sequence and learns to integrate the information of the question-answer context, as the resolution of the conversational sluice tends to repeat the structure of the antecedent of both the question and answer.

\paragraph{Inspection of model output}
Table \ref{tab:model_pred} present examples of conversational sluices from the test set along with the resolutions generated by our baselines as well as a gold annotated resolution.
From the examples, we can observe that the {\sc LSTM-seq2seq} often produces more nonsensical and less grammatically correct sentences, e.g. overusing question marks and inserting them in the middle of the sentences and it generally performs best when the input context and resolutions are short.%, 
The output of the {\sc Transformer} does improve upon the results of the {\sc LSTM-seq2seq}, producing more correct and coherent sentences,
however, the lack of pretraining compared to GPT-2, still results in less expressive sentences.
Most impressive are the results from the fine-tuned GPT-2 model. From its $r_1$ value we can see that almost 20\% of the instances, it actually generates a sluice resolution that our human judges ranked higher than the gold resolution. E.g., in the last sample generated by the GPT-2 model, demonstrates how it is able to incorporate all the information of the initial question $Q_1$, to a much higher degree than the what the annotator noted. The extensive pretraining does however allow the generated output to deviate a bit too much from the objective, as seen in the 3rd row.

\paragraph{Applying sluice resolutions in QA systems}
As mentioned in \S \ref{sec:intro}, the ability to resolve occurrences of ellipsis, either implicitly or explicitly, is important for question-answering system. With our gold annotated sluice resolutions, we replace instances of conversational sluices in the CoQA development set with their resolved counterparts, and evaluate the quality of the answers their baseline model provides.\footnote{Code for the pretrained CoQA baseline model is provided by  \url{https://github.com/stanfordnlp/coqa-baselines}}
In Figure \ref{tab:coqa_ex}, we see how the resolution of the conversational sluice leads to a much better answer, $A_{no-sluice}$, compared to the case where the model has to automatically draw the connection between \textit{`Why?'} and the context in $Q_1$ and $A_1$.
\begin{figure}[h]
    \centering
    \begin{tabularx}{10cm}{rl}
        $Q_1$:          & {\makecell[l]{What did Valetta think Mysie mustn't do?}} \\
        $A_1$:          & {Stay out after dark.} \\
        $Q_2$:          & {\makecell[l]{Why  [{\it does Valetta think that  Mysie} \\{\it shouldn't stay out after dark}]?}} \\
        \cmidrule(lr{1em}){2-2}
        $A_{no-sluice}$:   & {For fear she should cough.} \\
        $A_{sluice}$:& {no.} \\
        $A_{gold}$:     & {Fear she should cough.} \\
    \end{tabularx}
    \caption{A case where resolving the sluice in the an instance of the CoQA dataset improves the performance of QA system. $A_{no-sluice}$ is the answer generated when information contained in the bracket is included.}
    %\vspace{-3mm}
    \label{tab:coqa_ex}
\end{figure}
Of course injecting our annotations into the input at test time also biases the input data, making it less similar to the training data, and for this reason resolving sluices this way did not lead to significant improvements on average. %Resolving these occurrences will however not guarantee better performance if evaluated on models trained on data containing sluices, as a slight bias in introduced where questions in interconnected conversations now are stated more explicitly, differing from how we normally carry out more natural conversations.

% While state-of-the-art QA system are able to implicitly resolve...

\section{Conclusion}
This paper addresses the challenge of resolving occurrences of conversational sluices; that is, correctly identifying the antecedent of a bare {\it wh}-fronted ellipsis in a dialogue setting. We frame the task as a language generation task, where we %given a QA-context and a follow-up one-word question 
seek to generate the elided material. To this end, we crowd-sourced a new dataset of conversational sluices. We evaluate the performance of encoder-decoder architectures and language models on this data and show that human judges favour the resolutions generated by GPT-2, fine-tuned on our crowd-sourced annotations. Interestingly, resolutions rival the quality of human annotations. %We also demonstrate that resolving sluices in conversational question-answering datasets, can lead to improved performance of QA models.

\bibliography{aaai}
\bibliographystyle{aaai}
\begin{tabular}{rl}
\end{tabular}

\end{document}